\documentclass{article}
\usepackage{microtype}
\usepackage{graphicx}
\usepackage{subfigure}
\usepackage{booktabs} % for professional tables
\usepackage{amssymb}

\expandafter\def\csname ver@subfig.sty\endcsname{}
\usepackage{svg}
\usepackage{hyperref}

\usepackage{amsmath}
\usepackage{todonotes}
\usepackage{graphicx}
\usepackage{lipsum}
\usepackage{hyperref}

\usepackage{natbib}

\usepackage[hyperref]{utils/icml2021}
\usepackage{times}
\usepackage{latexsym}

\usepackage{caption}
\aclfinalcopy % Uncomment this line for the final submission
\title{Ask Before You Act \\ \it{Generalising to Novel Environments by Asking Questions}}

\author{Ross Murphy \\
  \texttt{\small ross.murphy.20} \\
    \texttt{\small @ucl.ac.uk} \\\And
  Sergey Mosesov \\
  \texttt{\small sergey.mosesov.13} \\
  \texttt{\small  @ucl.ac.uk} \\\And
  Javier Leguina Peral \\
  \texttt{\small javier.peral.20} \\
  \texttt{\small  @ucl.ac.uk} \\\And
  Thymo ter Doest \\
  \texttt{\small thymo.ter.doest.20} \\
    \texttt{\small  @ucl.ac.uk}} 
\begin{document}
\maketitle
\begin{abstract}
Solving temporally-extended tasks is a challenge for most reinforcement learning (RL) algorithms \cite{chelsea_finn}. We investigate the ability of an RL agent to learn to ask natural language questions as a tool to understand its environment and achieve greater generalisation performance in novel, temporally-extended environments. We do this by endowing this agent with the ability of asking ``yes-no" questions to an all-knowing Oracle. This allows the agent to obtain guidance regarding the task at hand, while limiting the access to new information. To study the emergence of such natural language questions in the context of temporally-extended tasks we first train our agent in a Mini-Grid environment. We then transfer the trained agent to a different, harder environment. We observe a significant increase in generalisation performance compared to a baseline agent unable to ask questions. Through grounding its understanding of natural language in its environment, the agent can reason about the dynamics of its environment to the point that it can ask new, relevant questions when deployed in a novel environment. 
\end{abstract}

\section{Introduction}
\label{intro}

Deep Reinforcement Learning is a powerful framework for sequential decision making that has propelled agents to surpass human performance in many settings \cite{hado_silver_ddqn, incentivise_exploration_abbeel_levine}. However, these methods suffer from many issues, including low sample efficiency, credit-assignment problem and struggle with hierarchical temporally-extended tasks \cite{minsky_61,sutton2018reinforcement}. Humans on the other hand, are able to quickly generalise to new environments by abstracting the structural similarities among various tasks \cite{spelke_kinzler}. This indicates humans learn in a different manner to machines \cite{human_atari_why}. This generalisation capability of humans can be attributed to their possessing a set of priors about the world they inhabit, and these priors can be grounded in natural language. \cite{spelke_kinzler, luketina}.

Furthermore, it has been hypothesised that, due to the need to encode information in the hierarchical and logical structure imposed by syntactically correct forms of expression, formal and natural languages allow us to encode abstractions and generalise \cite{gopmel}. We extend this reasoning and we question whether jointly learning to navigate an environment together with the language required to express your knowledge about it leads to a deeper understanding of the world. In other words, by forcing an agent to encode its knowledge of the world in a way amenable to expression through natural language, we argue that such generalisation capabilities will emerge naturally.

A wide body of research exists which seeks to improve generalisation in RL using natural language. Examples abound of approaches which jointly train an agent on a language goal and an environment, where the overarching concept is that the agent learns to encode information on the structure of the world  from natural language corpora (see \cite{luketina} and references therein). The idea being that the comprehension of language enables the agent to apply familiar heuristics to unfamiliar situations \cite{grounding_language_learning}.

However, the majority of these approaches use bespoke, human-composed corpora. It is difficult to see how this approach can scale in real-world applications. Instead, we advocate for giving agents the ability to ask natural language questions to an omniscient Oracle. By sampling from a vocabulary of relevant words, the agent must learn to generate syntactically correct questions parseable by the Oracle. Said Oracle then provides binary answers, to avoid the transmission of more information than desired. The agent then conditions its policy on the question-answer pair as well as an embedding of the history of previous state-question-answer-actions. To our knowledge, this is the first time an agent has been jointly trained to solve a task and do so by asking questions about its environment.

We wish to explore the following research questions:

\begin{enumerate}
\vspace{-0.2cm}
\setlength\itemsep{0em}
    \item Can an RL agent learn an effective joint policy over natural language questions and actions in its environment? 
    \item Will this joint policy lead to question asking behaviour which extracts information that is useful in solving the environment? 
    \item Can an RL agent generalise more effectively to novel environments through question-asking?
\end{enumerate}

% We frame our model through the lens of transfer-learning. We initially jointly train our agent on one environment, before dropping it into an unseen environment and assessing its ability to generalise in the new setting. We compare our model to a Baselineagent which is not endowed with this ability to converse in natural language with an Oracle. Our agent achieves far superior generalisation performance across a range of MiniGrid Environments (refer to hopefully nice looking experiments here...).

\section{Related Work}
\label{related}
\subsection{Natural Language Annotations}
An example of improving the generalisation performance of an agent by augmenting it with language understanding can be seen in ``RTFM" by \cite{rtfm}. The agent is jointly trained on both language and environment goals, as it is passed a document which describes the dynamics of its environment in natural language. This paper uses human-generated templates, and then ``procedurally" generates both the environment dynamics and descriptions of said dynamics by cycling through the possible combinations. With random ordering, the ``number of unique documents exceeds 15 billion", and this serves to ensure the agent cannot just memorise an environment - description mapping. The quantity of potential documents is outstanding, however the model still relies on human-composed text at its heart.

A similar approach can be seen in \cite{learn_to_win_silver} where the authors show an agent can achieve greater performance in the game of ``Civilization II" when it is jointly trained on playing the game and parsing the actual instruction manual for the game (which was naturally written by a human). \cite{branavan-etal-2009-reinforcement} present a RL method of learning a mapping from human-composed natural language instructions into sequences of executable actions, and show that (for well defined reward functions) this mapping can be learned despite having few if any labelled training examples (the agent learns solely via policy gradient methods backpropagating the reward). Our model has a very similar update method.

\subsection{Transfer-Learning}
A transfer-learning approach to the same problem can be seen in \cite{narasimhan-jaakkola}. The authors first ground an agent's understanding of natural language to the dynamics of an environment by passing the agent textual descriptions of the environment it was trained in. Greater generalisation performance is seen when subsequently training the agent in unseen environments with accompanying textual descriptions. A very similar approach is seen in \cite{harrison_transfer_learning} where the authors train an encoder-decoder network to learn associations between natural language descriptions and state/action information. This learned model then guides an RL agent's exploration in unseen environments, making it more effective at learning therein. 

While these papers both use natural language to improve the generalisation capabilities of RL agents, they differ from our model in that they both still rely on human-generated textual descriptions, rather than letting the agent generate its own Natural Language Corpus (NLC) through conversing with an Oracle.

\subsection{Meta-Learning}
\cite{grounding_policies_meta_learning_abbeel_levine} frame a similar problem as one of meta-learning, where the RL agent goes through a meta-training procedure, in order to learn a procedure, to enable it to adapt to new tasks during meta-test time. In this paper, the agent learns to guide its policy via ``iterative natural language corrections". This paper aligns closely with our model in that the temporal nature of the passing of natural language information to the agent is iterative, and the NLC is generated as the agent explores the new environment. This contrasts with the approaches seen in the ``Natural Language Annotations" section, where the NLC is passed to the agent in-one-go before it even begins in its new environment. In this paper the agent is continuously ``corrected" after each action, by receiving instructions such as ``Move closer to the pink block". Our model has a similar setup in that the agent receives an Answer to a Question it asks, after each action. Our model is similar, but differs markedly in two key ways: \begin{enumerate}
\vspace{-0.2cm}
\setlength\itemsep{0em}
    \item In our model, the onus is on the agent to ask about the pertinent information. The agent cannot just learn to receive natural language information, in our model it must learn to \textbf{ask} the right questions.
    \item In our model, the agent receives information from an Oracle which can only respond ``True" or ``False". This constrains the amount of ``extra" information the agent receives on each time-step to the the truth value of a query it itself has generated. 
\end{enumerate}

\subsection{Learning by Asking (LBA)}
The idea of Learning-by-Asking comes from \cite{lba}. The paper raises the key point that most visual-recognition models in the field of Computer Vision are 'passive' in that they are trained on a fixed corpus of data curated by humans, e.g. \cite{VQA}. The authors propose Learning-by-Asking (LBA), which endows an agent with the agency to decide which questions it needs to ask of an Oracle. The authors find their model is both more sample efficient and better at generalising to novel test-time distributions on the CLEVR dataset \cite{clevr_original}. Again this has many parallels with the setting of real-world interactive learning. This element of agency is the crucial feature which we carry over and apply to the problem of ``Language-Assisted RL" \cite{luketina}.

The architecture of our Q-A RNN ties in with a related branch of research, 'Visual Question Generation' (VQG) \cite{vqg}, a recent proposal as an alternative to image-captioning, in that we explicitly encourage (through the agent's loss) that the agent does ask Questions about items which do exist in the environment. We extend on this however and encourage our questions to not just be \textbf{relevant}, but also \textbf{informative}. We reinforce this by also explicitly including an entropy term in our loss function. Examples of using uncertainty measures to train networks can be seen in \cite{entropy}.

\section{Preliminaries}

\subsection{Reinforcement Learning Framework}
\begin{figure}
    \centering
    \includegraphics[trim={0.25cm 0cm 0.25cm 0cm}, clip, width=\linewidth]{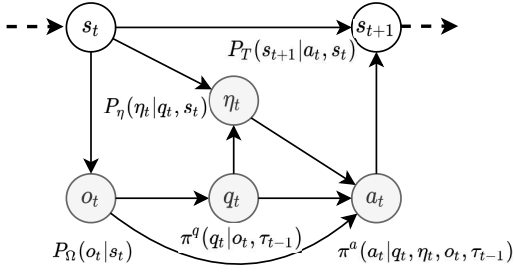}
    \caption{Graphical Model of POMDP. Dark-shading denotes observed variables, no-shading denotes latent variables.}
    \label{fig:graphical model}
\end{figure}

Our RL setting can be formalised as an \textit{augmented} Partially Observable Markov Decision Process (POMDP) \cite{sutton2018reinforcement} (see Figure \ref{fig:graphical model}) defined by the tuple $(\mathcal{S}, \mathcal{O}, \mathcal{Q}, \eta , \mathcal{A}, P_{\Omega}, P_{\eta}, P_T, R, \gamma)$, with $\mathcal{S}$ as the set of states, $\mathcal{O}$ as the set of observations, $\mathcal{Q}$ as the set of questions, $\eta$ as the set of Oracle answers, $\mathcal{A}$ the set of actions, $P_{\Omega}: \mathcal{S} \times \mathcal{O} \rightarrow  [0,1]$ as the conditional observation probability function,  $P_{\eta}: \mathcal{S} \times \mathcal{Q} \times \eta \rightarrow  [0,1]$ as the answer probability function, $P_T: \mathcal{S} \times \mathcal{A} \times \mathcal{S} \rightarrow  [0,1]$ as the transition probability function, $R: \mathcal{S} \times \mathcal{A} \rightarrow \mathbb{R}$ as the reward function, and $\gamma$ as the discount factor. 

We want to find a policy over actions, $\pi^a: \mathcal{A} \times \mathcal{O} \times  \mathcal{Q} \times \eta \to [0,1]$, that maximises the expected discounted return $\mathbb{E}_{s_0, a_0,...}[\sum^{\infty}_{t=0}\gamma^t r_{t+1}]$, where $a_t \sim \pi^a(a_t|o_t, q_t, \eta_t, \tau_{t-1})$, $s_t \sim P_T(s_{t+1}|s_t,a_t)$. Note that the policy over actions is conditioned on the current observation, question and answer, as well as $\tau_{t-1}$, the history of past observations, question, answers and actions $\tau_{t-1} = (o_{:t-1}, q_{:t-1}, \eta_{:t-1}, a_{:t-1})$ up to time $t-1$. 

The main hypothesis of this paper is that, by introducing an intermediate policy over questions, $\pi^q: \mathcal{O} \times  \mathcal{Q} \to [0,1]$, $q_t \sim \pi^q(q_t|o_t,\tau_{t-1})$ we can query information from an Oracle in the form of answers $\eta_t$. We argue that the resulting policy $\pi_\phi^a$ will perform better than a baseline without such a question-asking module, not only due to having access to more information about the environment, but also due to our hypothesis that by forcing an agent to encode its knowledge of the world in a way amenable to expression through natural language, further generalisation capabilities will emerge naturally. 

Note that, although expressed as a probability distribution the answer probability $P_\eta(\eta_t | q_t, s_t)$ is, in general, a deterministic function of the state and the question, given that questions are answered truthfully by the Oracle.

\section{Oracle} \label{sec: oracle}
The Oracle has full observability over the environment and is able to answer \texttt{True} or \texttt{False} to questions in relation to each object in the environment. The questions are posed as predicates such as ``\textit{red door is closed}" and ``\textit{green goal is north}".

\begin{figure}
    \centering
    \includegraphics[width=\linewidth]{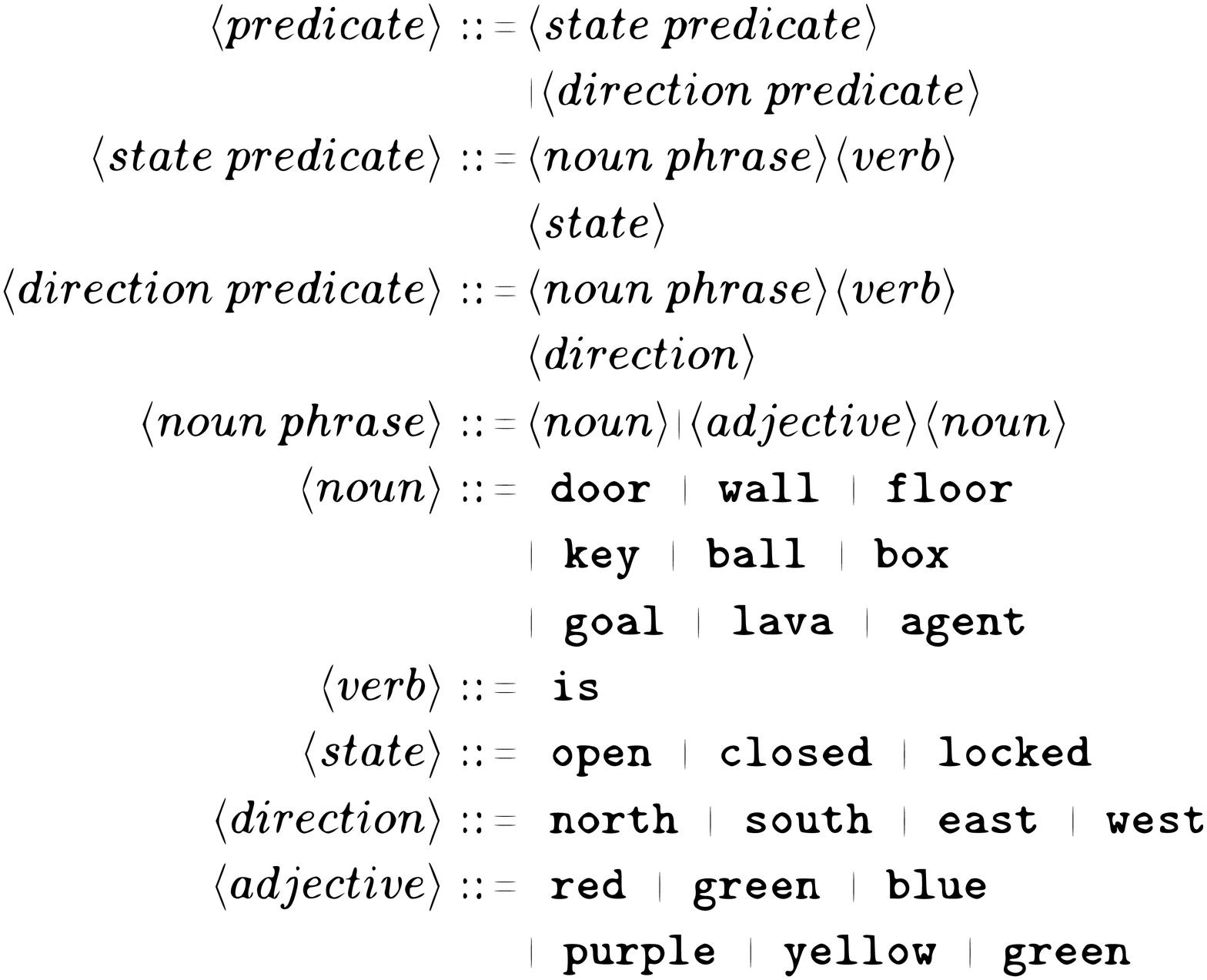}
    \caption{BNF Grammar}
    \label{fig:bnf}
\end{figure}

The Oracle is able to parse two types of predicates: 
\begin{itemize}

\vspace{-0.2cm}
\setlength\itemsep{0em}
    \item Direction predicates: the relative position of an object to the agent.
    \item State predicates: the current state of an object in the environment.
\end{itemize}  A predicate is well posed if it adheres to the BNF grammar in Figure \ref{fig:bnf}. The predicates can take one of four values  which defines $P_\eta(\eta_t | q_t, s_t)$:
\begin{itemize}
\vspace{-0.2cm}
\setlength\itemsep{0em}
    \item \texttt{True}: the predicate is well posed, uniquely identifies an object present in the environment and correctly describes the state or direction of the object; the Oracle returns $\eta_t = [1, 1]$ and a reward $r^q_t = 0.2$,
    \item \texttt{False}: the predicate is well posed, uniquely identifies an object present in the environment but incorrectly describes the state or direction of the object; the Oracle returns $\eta_t = [0, 0]$ and a reward $r^q_t = 0.2$,
    \item \texttt{Undefined}: the predicate is well posed, but does not uniquely identify and object in the environment and hence has no truth value; the Oracle returns $\eta_t = [0,1]$ and reward $r^q_t = 0$. 
    \item \texttt{Syntax error}: the predicate is not well posed, i.e. it does not adhere to the BNF grammar. The Oracle returns $\eta_t = [1,0]$ and rewards the agent with $r^q_t = -0.2$.
\end{itemize}

Note that the existence of an object in the current environment is not limited to the agent's perspective. In other words, the agent can ask questions related to objects not currently present in its observable neighbourhood. Through this, we aim to give the agent the ability to, potentially, learn and exploit a spatio-temporal knowledge representation of the objects it encounters by virtue of its ability to use memory. By asking questions about entities not currently in its sight, the agent could learn to associate the locality of said objects with a perception of time, insofar as objects which are not currently seen --but whose mention leads to an answer from the Oracle that indicates their existence-- either have been seen \textit{somewhere before} or will, potentially, be seen \textit{somewhere in the future}. Although the arousal of such behaviours is certainly unlikely at this stage, this does not preclude us from endowing the agent with the capability to do so, if only desideratively. 

\section{Ask Before You Act}

\begin{figure*}
    \centering
    \includegraphics[width=0.83\textwidth]{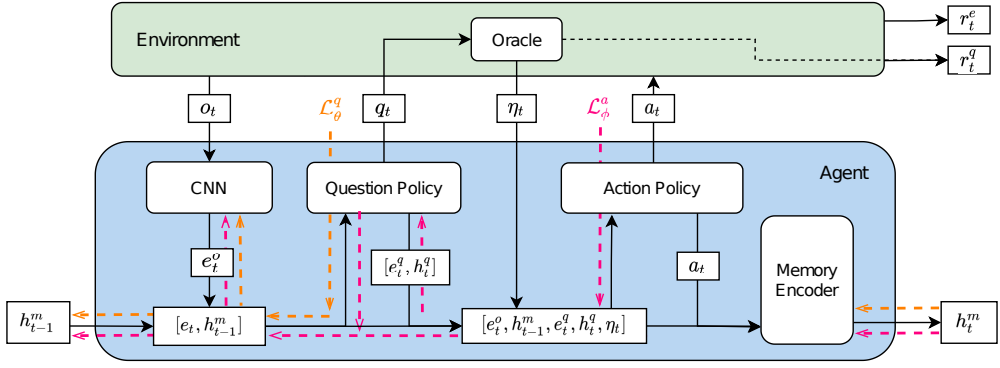}
    \caption{Model architecture with gradient flow.}
    \label{fig:model}
\end{figure*}

We propose an architecture that leverages a dialogue between the agent and an Oracle to guide the agent’s action policy. Our proposed model consists of a convolutional neural network \cite{lenet}, a memory encoder, and two policies: a neural network question-asking policy parametrised by $\theta$, $q_t \sim \pi^q_{\theta}(q_t|e^o_t,h_{t-1}^m)$, and a neural network action policy parametrised by $\phi$, $\pi^a_{\phi}(a_t|e^o_t, e^q_t, \eta_t, h^m_{t-1})$ (Figure \ref{fig:model}). Note that the conditioning variables of the policies are no longer the general random variables $o_t$, $q_t$, $\tau_{t-1}$ but a neural embedding of such variables $e^o_t$, $e^q_t$, $h^m_{t-1}$ respectively. This is done to retain only the most relevant features of the inputs.  

At every time-step $t$ the agent receives a partial observation of the environment $o_t$, which is encoded via a convolutional neural network parametrised by $\nu$ to a latent lower-dimensional space $e_t^o = f_\nu(o_t)$. Together with this observation, the agent carries forward the hidden state of an LSTM network (the memory module, Figure \ref{fig:model}) \cite{lstm}, whose purpose is to encode the history of observations, questions, answers and actions up to time step $t-1$:
\begin{subequations}
\begin{equation}
    h_{0}^m \sim \mathcal{N}(0, I) 
\end{equation}
\begin{equation}
    h_{t}^m = f_\mu(h_{t-1}^m, e^o_{t-1}, e^q_{t-1}, \eta_{t-1}, a_{t-1}),
\end{equation}
\end{subequations}
where $f_\mu$ is an LSTM network parametrised by $\mu$. The concatenated vector $[e_t^o, h_{t-1}^m]$, is then passed to the question policy $\pi^q_{\theta}(q_t|e_t^o,h_{t-1}^m)$.

\subsection{Question Policy}
The agent's question policy is responsible for generating relevant questions given the current observation and the hidden state of the memory embedding. The inclusion of the trajectory embedding allows the agent to ask questions with a larger time horizon. By retaining information from previous observations, question-answer pairs and actions, the agent need not ask the same questions when in similar states, thus being able to explore more efficiently and generalise better.

More specifically, the question policy consists of a pre-trained generative language model. The question $q_t$ distribution is modelled according to
\begin{align} 
\pi^q_{\theta}(q_t|e_t^o,h_{t-1}^m) & = \pi^q_{\theta}\left(w_{1:N}|e_t^o,h_{t-1}^m\right) \\
& = \prod_{i=1}^{N} p_{\theta}\left(w_{t} \mid w_{1:t-1}, e_t^o,h_{t-1}^m\right) \label{eq: joint tokens}
\end{align}
with $w_i$ as the individual tokens.
The language model consists of a word embedding layer, LSTM layer and a fully connected layer. The word embedding layer is a mapping from the vocabulary (which consists of all valid tokens in the BNF grammar) to a latent space using $e^{w_i} = g_{\theta}(w_i)$.
% $e_t^{w_i}$

The model is pre-trained on a corpus containing all possible phrases which adhere to the BNF grammar Figure \ref{fig:bnf}.
We pre-train the model in a supervised manner on this corpus using cross-entropy loss. The resulting pre-trained model is then used to initialise the question policy $\pi^q_\theta$ which is then fine-tuned using the model updates of the agent (see Section \ref{sec: update}) with the word embedding layer fixed.

% \begin{equation}
% \mathcal{L}_{p_{\theta}}(\hat{w})=-\frac{1}{N} \sum_{i} \hat{y}_{\hat{w}}^{T} \log p_{\theta}\left(\hat{w}\mid w_{i-1}, \ldots, w_{1}\right)
% \end{equation}

% $h^{q_0}_t = [e_t^o, h_{t-1}^m]$,

At every time step $t$ the question policy $\pi^q_{\theta}(q_t|e^o_t,h_{t-1}^m)$ takes as its initial hidden state the concatenated $h_t^{w_0} = [e_t^o, h_{t-1}^m]$, and as input the embedding vector corresponding to the $<$sos$>$ token $w_0 = g_{\theta}(<$sos$>)$. We denote $h_t^{w_0}$ as the initial hidden state of the $\pi^q_{\theta}$ at the (environment) time step $t$. The hidden state $h_t^{w_i}$ is computed recursively as
\begin{align} 
h_t^{w_i} = f_\theta(w_i, h_t^{w_{i-1}})
\end{align}
where $f_\theta(\cdot)$ is a single pass through the embedding layer and LSTM cell.
% The output from each LSTM cell is passed through a single fully connected layer and softmax layer over the vocabulary size.
We then pass the hidden state $h_t^{w_i}$ through a single fully connected layer and softmax layer, after which we sample $w_i$ the $i$'th token of the question.
This process continues until the $<$eos$>$ token is sampled. Then the question $q_t$ is the concatenation of the sampled tokens without the $<$sos$>$ and $<$eos$>$ tokens $q_t = [w_1,\ldots, w_{N-1}]$.

The resulting question $q_t$ is passed to the Oracle, which returns a tuple $(\eta_t, r^q_t)$, where $\eta_t$ is the answer to said question and $r^q_t$ is a reward (see Section \ref{sec: oracle}). 

\subsection{Action Policy}

The resulting answer $\eta_t$ from the Oracle is concatenated with the last hidden state from the question policy $\pi^q_{\theta}$, denoted $h_t^{q}$ and to an embedding of the question $e^q_t$. The question embedding $e^q_t$ is computed by mean pooling the word embeddings $e_t^{w_i}$. Together with the embedding of the current observation $e^o_t$ and the memory hidden state $h^m_{t-1}$, this conforms the input to the action policy $\pi^a_{\phi}$ (see Figure \ref{fig:model}).

As mentioned before, the action policy $\pi^a_{\phi}$ consists of a neural network parametrised by $\phi$, with a softmax layer over the number of environment actions as its final layer. The action at each time step is sampled from this distribution $a_t \sim \pi_\phi^a(a_t|e^o_t, e^q_t, \eta_t, h^m_t)$. 

Finally, the observation embedding, question embedding, answer and action are passed as the input to the memory module, which encodes it and outputs the encoded hidden state for the next time step.

\subsection{Update}
\label{sec: update}
\paragraph{Action Loss.} The action policy is trained using a PPO style loss:
\begin{subequations}
\begin{equation}
    \mathcal{L}^a_t(\phi) = \left[ \mathcal{L}_t^{clip}(\phi) - c_1 \mathcal{L}_t^{vf}(\phi) + c_2 H[\pi^a_\phi] \right]
\end{equation}
\begin{equation}
    \mathcal{L}_t^{clip} = \hat{A}^{GAE}_t \min \left[r^e_t(\phi), \text{clip}(r^e_t(\phi), 1 \pm \epsilon) \right]
\end{equation} \vspace{0pt}
\begin{equation}
    \mathcal{L}_t^{vf} = \mathcal{L}_{H}\left(V_\phi(s_t) - V_t^{target} \right) = \mathcal{L}_{H} (\delta_t), 
\end{equation}
\end{subequations}
where $r^e_t(\phi)$ is the probability ratio
\begin{equation}
    r^e_t(\phi) = \frac{\pi_\phi^a(a_t|e^o_t, e^q_t, \eta_t, h^m_t)}{\pi_{\phi_{old}}^a(a_t|e^o_t, e^q_t, \eta_t, h^m_t)},
\end{equation}
so $r^e_t(\phi_{old}) = 1$; $\hat{A}^{GAE}_t$ is the generalised advantage function 
\begin{equation}
    \hat{A}^{GAE}_t = \sum_{l=0}^\infty (\gamma \lambda)^l \delta^V_{t+l}
\end{equation} 
with $\delta = r^e_t + \gamma V_{\phi}(s_{t+1}) - V_{\phi}(s_t) $ being the temporal-difference error \cite{sutton2018reinforcement} and $\mathcal{L}_{H_1}$ denotes the smooth L1 loss or Huber loss for $a=1$. \cite{huber}. Moreover, $\text{clip}(r^e_t(\phi), 1 \pm \epsilon)$ saturates the ratio $r^e_t(\phi)$ to be in the interval $[1 - \epsilon, 1 + \epsilon]$. $H[\pi_\phi^a]$ is an entropy regulariser which encourages higher entropy in the policy distribution, increasing the extent to which the agent explores the environment. The constants $c_1$ and $c_2$ are hyperparameters that weight the relative importance of the different losses (see Table \ref{tab:hyper}).

Although not shown in Figure \ref{fig:model}, the value estimate $V_\phi(s_t)$ is generated by a neural network built on top of the policy network, with the softmax layer replaced by a fully connected layer from the number of actions to a scalar. 

\paragraph{Question Loss.}
The loss applied to the question policy is a modified version of REINFORCE \cite{sutton2018reinforcement} with entropy regularisation: 
\begin{multline}
    \mathcal{L}^q_t(\theta) = (c_3 r^q_t + c_4 G_t ) \ln \pi^q_{\theta}(q_t|e_t^o,h_{t-1}^m) \\ 
    + c_5 H[\pi^q_{\theta}],
\end{multline}
where $c_3$, $c_4$, $c_5$ are hyperparameters \ref{tab:hyper} and $G_t$ is episode reward at time step $t$ \cite{sutton2018reinforcement}:
\begin{equation}
    G_t =\sum^T_{k=t+1} \gamma^{k-t-1} r^e_k.
\end{equation}
Note that this value can be computed since we only perform gradient steps at at the end of each episode.

The purpose of this update is twofold: through $r^q_t$ we enforce that the questions are syntactically correct and address objects present in the current environment (see Section \ref{sec: oracle}); through $G_t$ we increase the episode return associated to the questions which influenced the policy in such a way that the question policy maximises the expected return the environment. 

Note that the entropy term in the question loss leads to increased entropy in the policy distribution over words and thus encouraging the agent to ask more diverse questions. 

\paragraph{Backpropagation}
During the course of an episode, transitions are stored in a buffer. At the end of every episode these transitions are used to calculate the losses, which are combined into 
\begin{equation}
    \mathcal{L}(\phi, \theta) = \mathcal{L}^q_t(\theta) + \mathcal{L}^a_t(\phi).
\end{equation}
A single gradient step then is taken on $-\mathcal{L}(\phi, \theta)$ using ADAM as an optimiser with a fixed learning rate $\alpha$ (See Table \ref{tab:hyper}) \cite{adam}. The data buffer is emptied after each episode.

Finally, since the model trains end-to-end, through the above losses we also update the parameters $\nu$ of the CNN. The parameters $\mu$ of the memory LSTM are also updated with this loss, which is backpropagated through time until the first transition of the episode. See Figure \ref{fig:model} for a diagram of the gradient flow overlaid onto the model.

\begin{figure*}[t!]
    \centering
    \includegraphics[width=0.87\linewidth]{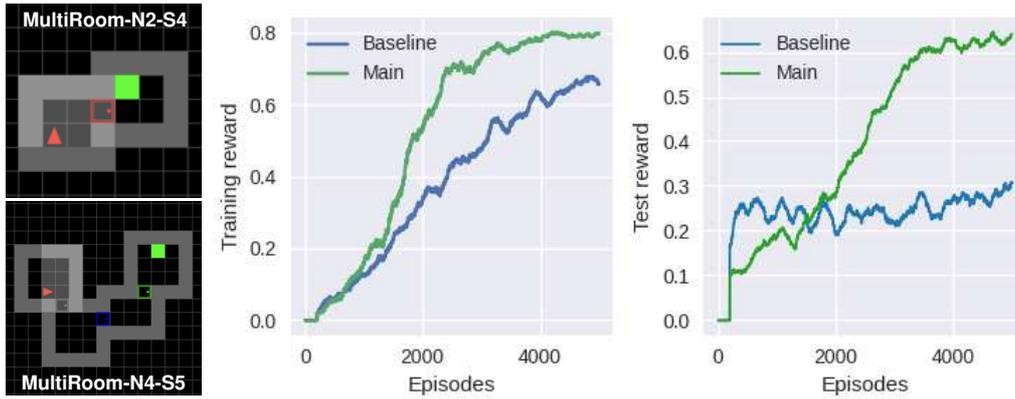}
    \caption{Top left: Sample of the training environment. Bottom left: Sample of the testing environment. Left graph: 100 episode moving average reward during training, averaged over 7 seeds. Right graph: 100 episode moving average reward during testing, averaged over 7 seeds.}
    \label{fig:generalisation results}
\end{figure*}

\section{Experiments}
For experiments we use the MiniGrid environment \cite{gym_minigrid}. This is a series of fast, lightweight grid world environments. More specifically, we use the so-called MultiRoom set of environments, in which an agent needs to traverse a set of rooms separated by doors in order to get to the model. These environments provide two main features which make them attractive for our purposes: first, they have a reduced set of objects present (walls, coloured doors and a goal) which allows for easy language acquisition; and they are procedurally generated, which can stop the agents from simply memorising the environment. 

For these environments, the reward is calculated as:
\begin{equation}
    r^e_t = \begin{cases}
    0 & \text{if} \quad t \neq T \\
    1 - \frac{9}{10} \frac{T}{T_{max}} & \text{otherwise}
    \end{cases}
\end{equation}
where $T_{max} = 20 S_{room}$, which depends on the maximum possible size of the rooms in the environment. Finally, all reported results are averaged over 7 different environment seeds.

\subsection{Baseline}

For the experiments we use a PPO baseline model \cite{ppo}, which also uses Generalised Advantage Estimation (GAE) \cite{gae}. To make the baseline comparable to our model, we also add a convolutional neural network and memory module with the same architectures as the main model. The only difference between the baseline and our main model is that the baseline is not augmented with a question-asking component. As above, the model is trained end-to-end, with a single gradient step done in batch for each episode using ADAM. The learning rate and other hyperparameters can be seen in Table \ref{tab:hyper}.

\subsection{Generalising to Unseen Environments}\label{generalisation}

\begin{table}[]
\caption{Table summarising the results obtained for generalisation (Figure \ref{fig:generalisation results}) and noisy Oracle (Figure \ref{fig:noisy results}) experiments. Results show mean $\pm$ std. dev. calculated over 7 environment seeds.}
\centering
\begin{tabular}{lll}\hline
Model     & Train         & Test          \\ \hline
Baseline  & 0.737 ± 0.114 & 0.202 ± 0.239 \\
\textbf{Main}      & \textbf{0.803 ± 0.011} & \textbf{0.711 ± 0.135} \\
Main (Rand Oracle) & N/A           & 0.565 ± 0.207 \\
FiLM      & 0.770 ± 0.067 & 0.504 ± 0.277 \\
FiLM (Rand Oracle) & N/A           & 0.258 ± 0.331 \\ \hline
\end{tabular}
\label{tab:results}
\end{table}

We train both the baseline and the main agent on the \texttt{MiniGrid-MultiRoom-N2-S4-v0} environment until convergence. This environment consists of two rooms with maximum size of $S_{room} = 4$ separated by a locked door. The agent must open and pass through the door in order to reach the goal located in the other room, as shown in Figure \ref{fig:generalisation results}. The layout of the environment is randomly generated at every episode so the agent cannot simply memorise a sequence of steps to the goal.

We subsequently deploy both agents on a novel environment the larger, and more complex \texttt{MiniGrid-MultiRoom-N4-S5-v0} environment. This environment comprises four rooms with maximum size of $S_{room} = 5$ separated by locked doors, with the goal located in the final room (Figure \ref{fig:model}). Both environments are procedurally generated with varying door colours and room locations. When testing the agents, we still perform gradient updates, as to allow the models to adapt to the new environment.  This follows the curriculum learning setting, evaluating the agents ability to transfer learn to a harder environment \cite{curriculum_learning}.

The only difference between the train and testing regimes lies in the reward given by the Oracle. During testing, the agent is only penalised for syntactically incorrect premises, but receives no reward otherwise i.e. we no longer include a positive reward for uniquely identifying an object in the environment.

We wish to evaluate the agent's ability to ground its understanding of natural language in the train environment, and carry over this knowledge to the unseen environment without further reinforcement for correctly identifying objects in the new environment. All other experiments follow the same methodology outlined above.

The mean rewards across all runs are recorded in Figure \ref{fig:generalisation results}. Full results can be seen in Table \ref{tab:results}. We observe that the main agent outperforms the baseline both during training and testing. These results indicate that the question asking module in concert with the Oracle's answers allow the main agent to generalise more effectively to the unseen environment. 

\begin{figure*}[]
    \centering
    \includegraphics[width=\linewidth]{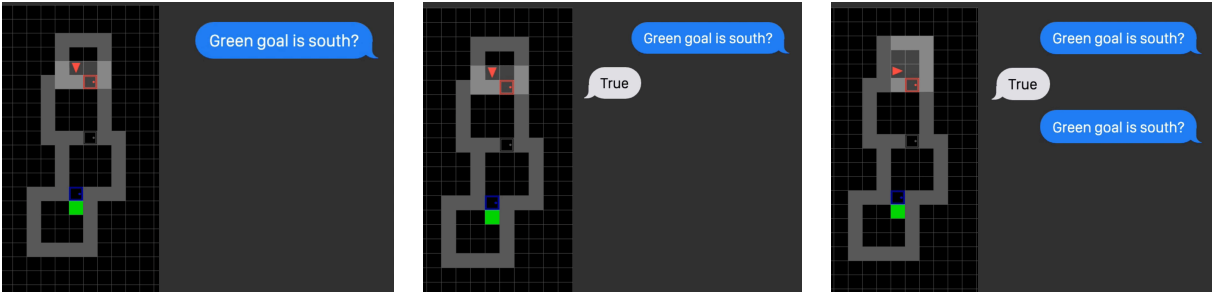}
    \caption{Qualitative demonstration of the agent behaviour in the test environment.}
    \label{fig:qualitative results}
\end{figure*}

\subsection{Qualitative Assessment}

When qualitatively assessing the dialogue generated between the agent and the Oracle, we noticed that the majority of the agent's questions referred to the ``\textit{green goal}". See Figure \ref{fig:qualitative results} for a qualitative demonstration of said behaviour. This is the expected behaviour, when one considers that the agent is penalised during training for asking about objects which are not present in its environment. Thus, the agent learns to ask about as the one object which is present across all MiniGrid environments and will incur in the smallest penalisation. 

We question whether this behaviour is a result of the language grounding or a result of the statistical phenomena outlined above, especially given that in some instances the agent does act logically following certain questions about the location of the green goal, whereas in others, we fail to see a significant correlation between the questions and the actions.

More specifically, in certain episodes we observed a directly interpretable correspondence between the question-answer pairs and actions. For instance, the agent may ask "\textit{green goal is south?}", receive the answer \texttt{True} and turn south.

However the agent would sometimes appears to ignore the question-answer pairs when taking the immediate action. Although  one would expect the influence of the question-answer pair to manifest itself immediately after receiving an answer, it is unclear whether the agent entirely disregards the new information or if the QA pairs are being encoded in the memory module and are acted upon at a later time step. Nevertheless, the promising quantitative results seem to indicate that the information is being used at some point.

\subsection{Ablations \& Extensions}

\subsubsection{FiLM}
Given the success of \cite{rtfm}, we also experimented with an alternative model architecture incorporating FiLM layers \cite{film}. Rather than passing the encoding of the QA pairs to the policy and value networks we instead condition the observation on the QA pairs using a FiLMed CNN network \cite{film}. The CNN network is extended with five additional Residual Convolution Layers (ResBlocks). The FiLM layer cosists of a fully connected linear layer which takes the encoding of the QA pairs as an input, and outputs the parameters of an affine-transformation which is applied to the activations of the ResBlocks.

\begin{figure}[t]
    \centering
    \includegraphics[trim={0cm 0cm 0cm 0cm}, clip, width=0.9\linewidth]{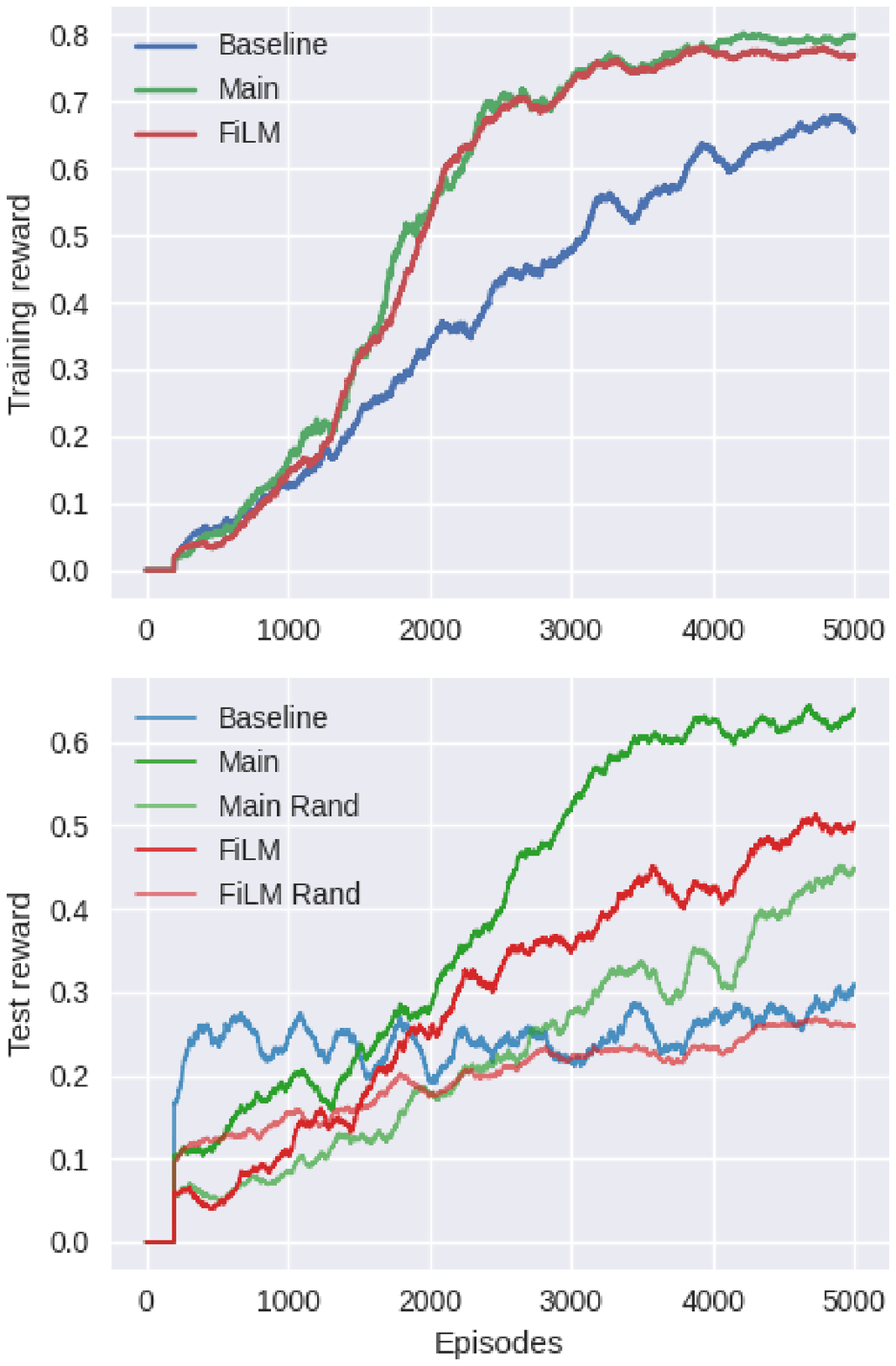}
    \caption{Top Plot: Training main, main agent, and FiLM agent on \texttt{MiniGrid-MultiRoom-N2-S4-v0}. Bottom Plot: Testing both main and main agent on \texttt{MiniGrid-MultiRoom-N4-S5-v0}.}
    \label{fig:noisy results}
\end{figure}

This addition did not significantly improve the performance of our model. We hypothesise this is because our GridWorld environments are not sufficiently visually complex, so the benefit of the FiLM architecture is slight. In other words, exploration of the environment via the visual domain seems an easier option than the exploration through the ``information domain", due to , perhaps, a not-too-reduced partial observability.  Nonetheless, we included this model in Figure \ref{fig:noisy results}, and we publish our corresponding code in our GitHub repository. \footnote{\href{https://github.com/ser-ge/ask_before_you_act}{\texttt{www.github.com/ser-ge/ask\_before\_you\_act}}}

\subsubsection{Random Oracle}

We again train our main, FiLM, and baseline agents on the \texttt{MiniGrid-MultiRoom-N2-S4-v0} and test them in the \texttt{MiniGrid-MultiRoom-N4-S5-v0}, as before. However, in this ablation, during testing we also test the agents with a ``confused" Oracle, which responds to the agent with random answers.

We compare how the test performance of the agents differs when the Oracle is responding with useful answers versus random noise. These results are shown in Figure \ref{fig:noisy results} and are also in Table \ref{tab:results}.

We note that both the main and the FiLM extended agents both see superior performance in the test environment when receiving correct answers from the Oracle, versus when they receive random answers. We interpret these results as indicating that the agents are succeeding, to some extent, in internalising the information which they receive by asking questions, in order to understand their novel environment.

\subsubsection{Mutual Information Regularisation}

Given the observed weak conditioning of the actions on the questions for some training instances, we considered how we could enforce a stronger conditioning between QA pairs and the subsequent actions. 

We formalise this objective by performing a maximisation of the mutual information between an action $a_t$ and its preceding question $q_t$
% \begin{align}\label{eq: mutual info}
%     I(a_t; q_t) & = H(a_t) - H(a_t | q_t)\\ \label{eq: mutual info 2}
%     & = \sum_{\mathcal{A}} p(a_t) \log p(a_t) \\ \label{eq: mutual info 3}
%     & - \sum_{\mathcal{A}} p(a_t|q_t) \log p(a_t|q_t) 
% \end{align} 

\begin{align}\label{eq: mutual info}
    I(a_t; q_t) & = H(a_t) - H(a_t | q_t)\\ \label{eq: mutual info 2}
    & = \sum_{\mathcal{A}}  p(a_t) \log p(a_t)- p(a_t|q_t) \log p(a_t|q_t) 
\end{align} 
where the other conditioning variables are implied for ease of notation. The mutual information as expressed in (\ref{eq: mutual info}) quantifies how much can be known about $a_t$ once $q_t$ is observed. Note that since the answer $\eta_t$ is a deterministic function of the question and the state given by the environment, we cannot treat it as a variable in our model, and so we make it implicit in our notation.

We see that the RHS of expression (\ref{eq: mutual info}) is simply the entropy of the action policy, as given by our model $H[\pi_\phi^a]$, quantity which is also used for the loss functions in Section \ref{sec: update}. However, we do not know the marginal probability of $a_t$, which is required for the LHS of expression (\ref{eq: mutual info 2}). 

To compute this quantity, we would want to marginalise $q_t$ over the joint probability of $q_t$ and $a_t$ 
\begin{equation} \label{eq: joint}
p(a_t) = \sum_{\mathcal{Q}} p(a_t, q_t) = \sum_{\mathcal{Q}}  p(a_t|q_t)p(q_t),
\end{equation}
where $p(q_t)$ can be estimated from the softmax distributions over tokens given by the LSTM question policy, and $p(a_t|q_t)$ is just a forward pass of the action policy. 

We can approximate $p(a_t)$ by performing $N$ number of forward passes on the question policy and computing for each question the joint probability over tokens as per expression (\ref{eq: joint tokens}). We then pass each of the questions to the Oracle, which returns $N$ different answers $\eta_t$. Given each of the $N$ QA pairs, we can now run the policy forward pass to obtain $N$ values of $p(a_t|q_t)$. By multiplying each $p(a_t|q_t)$ with its respective distribution over actions $p(a_t)$ and summing over the $N$ samples, we can obtain an estimate of $p(a_t)$. It remains to calculate the entropy of the resulting distribution as in (\ref{eq: mutual info 2}).

Finally, by incorporating the mutual information $I(a_t; q_t)$ into the loss of the model, we can force this term to be maximal, relative to the other terms in the loss function. In principle, this could improve the explainability of the actions in light of the questions.

We initially extended our model to include this feature, however the sampling process proved too computationally inefficient. As such, we excluded this feature from our final model.

\section{Conclusion}

We propose \textit{Ask Before You Act}, a problem setting in which an RL agent must learn a joint policy over actions and natural language questions posed to an all knowing Oracle, to obtain information that would assist it in solving its environment. We investigate the ability of the main agent to successfully query useful information by evaluating its ability to generalise to unseen harder environments having grounded its natural language understanding in the training environment. The main agent performs well in this curriculum setting, learning a joint policy over action and questions which significantly outperforms the baseline in the harder environment.  In order to evaluate the usefulness of the Oracle's answers to the agent we performed ablations with a random Oracle in the test environment. We observed a material reduction in performance with a random Oracle, indicating that the agent successfully leverages the ability to query information through natural language. 

Despite promising quantitative results the qualitative results were inconclusive. We observed that at times the agents' behaviour did not correspond directly to the dialogue with the Oracle; or that the agent would fixate on asking questions about the same object (i.e. green goal). Further work may focus on more challenging generalisations settings with no object permanence across environments. In addition one may wish to conduct experiments with a tighter coupling of action policy to the questions through mutual information regularisation which would allow for a more robust assessment of the role of dialogue in solving the environment.

\bibliography{biblio}

\begin{thebibliography}{30}
\providecommand{\natexlab}[1]{#1}
\providecommand{\url}[1]{\texttt{#1}}
\expandafter\ifx\csname urlstyle\endcsname\relax
  \providecommand{\doi}[1]{doi: #1}\else
  \providecommand{\doi}{doi: \begingroup \urlstyle{rm}\Url}\fi

\bibitem[Jiang et~al.(2019)Jiang, Gu, Murphy, and Finn]{chelsea_finn}
Yiding Jiang, Shixiang Gu, Kevin Murphy, and Chelsea Finn.
\newblock Language as an abstraction for hierarchical deep reinforcement
  learning.
\newblock \emph{CoRR}, abs/1906.07343, 2019.
\newblock URL \url{http://arxiv.org/abs/1906.07343}.

\bibitem[van Hasselt et~al.(2015)van Hasselt, Guez, and
  Silver]{hado_silver_ddqn}
Hado van Hasselt, Arthur Guez, and David Silver.
\newblock Deep reinforcement learning with double q-learning.
\newblock \emph{CoRR}, abs/1509.06461, 2015.
\newblock URL \url{http://arxiv.org/abs/1509.06461}.

\bibitem[Stadie et~al.(2015)Stadie, Levine, and
  Abbeel]{incentivise_exploration_abbeel_levine}
Bradly~C. Stadie, Sergey Levine, and Pieter Abbeel.
\newblock Incentivizing exploration in reinforcement learning with deep
  predictive models.
\newblock \emph{CoRR}, abs/1507.00814, 2015.
\newblock URL \url{http://arxiv.org/abs/1507.00814}.

\bibitem[Minsky(1961)]{minsky_61}
Marvin Minsky.
\newblock steps.pdf.
\newblock \url{https://courses.csail.mit.edu/6.803/pdf/steps.pdf}, 1961.
\newblock (Accessed on 05/31/2021).

\bibitem[Sutton and Barto(2018)]{sutton2018reinforcement}
Richard~S Sutton and Andrew~G Barto.
\newblock \emph{Reinforcement learning: An introduction}.
\newblock MIT press, 2018.

\bibitem[Spelke and Kinzler(2007)]{spelke_kinzler}
Elizabeth~S. Spelke and Katherine~D. Kinzler.
\newblock Core knowledge.
\newblock \emph{Developmental Science}, 10\penalty0 (1):\penalty0 89--96, 2007.
\newblock \doi{https://doi.org/10.1111/j.1467-7687.2007.00569.x}.
\newblock URL
  \url{https://onlinelibrary.wiley.com/doi/abs/10.1111/j.1467-7687.2007.00569.x}.

\bibitem[Tsividis et~al.(2017)Tsividis, Pouncy, Xu, Tenenbaum, and
  Gershman]{human_atari_why}
Pedro Tsividis, Thomas Pouncy, Jaqueline Xu, Joshua Tenenbaum, and Samuel
  Gershman.
\newblock Human learning in atari, 2017.
\newblock URL
  \url{https://www.aaai.org/ocs/index.php/SSS/SSS17/paper/view/15280}.

\bibitem[Luketina et~al.(2019)Luketina, Nardelli, Farquhar, Foerster, Andreas,
  Grefenstette, Whiteson, and Rockt{\"{a}}schel]{luketina}
Jelena Luketina, Nantas Nardelli, Gregory Farquhar, Jakob~N. Foerster, Jacob
  Andreas, Edward Grefenstette, Shimon Whiteson, and Tim Rockt{\"{a}}schel.
\newblock A survey of reinforcement learning informed by natural language.
\newblock \emph{CoRR}, abs/1906.03926, 2019.
\newblock URL \url{http://arxiv.org/abs/1906.03926}.

\bibitem[Gopnik and Meltzoff(1987)]{gopmel}
Gopnik and Meltzoff.
\newblock The development of categorization in the second year and its relation
  to other cognitive and linguistic developments on jstor.
\newblock \url{https://www.jstor.org/stable/1130692?seq=1}, 1987.
\newblock (Accessed on 05/30/2021).

\bibitem[Hermann et~al.(2017)Hermann, Hill, Green, Wang, Faulkner, Soyer,
  Szepesvari, Czarnecki, Jaderberg, Teplyashin, Wainwright, Apps, Hassabis, and
  Blunsom]{grounding_language_learning}
Karl~Moritz Hermann, Felix Hill, Simon Green, Fumin Wang, Ryan Faulkner, Hubert
  Soyer, David Szepesvari, Wojciech~Marian Czarnecki, Max Jaderberg, Denis
  Teplyashin, Marcus Wainwright, Chris Apps, Demis Hassabis, and Phil Blunsom.
\newblock Grounded language learning in a simulated 3d world.
\newblock \emph{CoRR}, abs/1706.06551, 2017.
\newblock URL \url{http://arxiv.org/abs/1706.06551}.

\bibitem[Zhong et~al.(2019)Zhong, Rockt{\"{a}}schel, and Grefenstette]{rtfm}
Victor Zhong, Tim Rockt{\"{a}}schel, and Edward Grefenstette.
\newblock {RTFM:} generalising to novel environment dynamics via reading.
\newblock \emph{CoRR}, abs/1910.08210, 2019.
\newblock URL \url{http://arxiv.org/abs/1910.08210}.

\bibitem[Branavan et~al.(2014)Branavan, Silver, and
  Barzilay]{learn_to_win_silver}
S.~R.~K. Branavan, David Silver, and Regina Barzilay.
\newblock Learning to win by reading manuals in a monte-carlo framework.
\newblock \emph{CoRR}, abs/1401.5390, 2014.
\newblock URL \url{http://arxiv.org/abs/1401.5390}.

\bibitem[Branavan et~al.(2009)Branavan, Chen, Zettlemoyer, and
  Barzilay]{branavan-etal-2009-reinforcement}
S.R.K. Branavan, Harr Chen, Luke Zettlemoyer, and Regina Barzilay.
\newblock Reinforcement learning for mapping instructions to actions.
\newblock In \emph{Proceedings of the Joint Conference of the 47th Annual
  Meeting of the {ACL} and the 4th International Joint Conference on Natural
  Language Processing of the {AFNLP}}, pages 82--90, Suntec, Singapore, aug
  2009. Association for Computational Linguistics.
\newblock URL \url{https://www.aclweb.org/anthology/P09-1010}.

\bibitem[Narasimhan et~al.(2017)Narasimhan, Barzilay, and
  Jaakkola]{narasimhan-jaakkola}
Karthik Narasimhan, Regina Barzilay, and Tommi~S. Jaakkola.
\newblock Deep transfer in reinforcement learning by language grounding.
\newblock \emph{CoRR}, abs/1708.00133, 2017.
\newblock URL \url{http://arxiv.org/abs/1708.00133}.

\bibitem[Harrison et~al.(2017)Harrison, Ehsan, and
  Riedl]{harrison_transfer_learning}
Brent Harrison, Upol Ehsan, and Mark~O. Riedl.
\newblock Guiding reinforcement learning exploration using natural language.
\newblock \emph{CoRR}, abs/1707.08616, 2017.
\newblock URL \url{http://arxiv.org/abs/1707.08616}.

\bibitem[Co{-}Reyes et~al.(2018)Co{-}Reyes, Gupta, Sanjeev, Altieri, DeNero,
  Abbeel, and Levine]{grounding_policies_meta_learning_abbeel_levine}
John~D. Co{-}Reyes, Abhishek Gupta, Suvansh Sanjeev, Nick Altieri, John DeNero,
  Pieter Abbeel, and Sergey Levine.
\newblock Guiding policies with language via meta-learning.
\newblock \emph{CoRR}, abs/1811.07882, 2018.
\newblock URL \url{http://arxiv.org/abs/1811.07882}.

\bibitem[Misra et~al.(2017)Misra, Girshick, Fergus, Hebert, Gupta, and van~der
  Maaten]{lba}
Ishan Misra, Ross~B. Girshick, Rob Fergus, Martial Hebert, Abhinav Gupta, and
  Laurens van~der Maaten.
\newblock Learning by asking questions.
\newblock \emph{CoRR}, abs/1712.01238, 2017.
\newblock URL \url{http://arxiv.org/abs/1712.01238}.

\bibitem[Antol et~al.(2015)Antol, Agrawal, Lu, Mitchell, Batra, Zitnick, and
  Parikh]{VQA}
Stanislaw Antol, Aishwarya Agrawal, Jiasen Lu, Margaret Mitchell, Dhruv Batra,
  C.~Lawrence Zitnick, and Devi Parikh.
\newblock {VQA:} visual question answering.
\newblock \emph{CoRR}, abs/1505.00468, 2015.
\newblock URL \url{http://arxiv.org/abs/1505.00468}.

\bibitem[Johnson et~al.(2016)Johnson, Hariharan, van~der Maaten, Fei{-}Fei,
  Zitnick, and Girshick]{clevr_original}
Justin Johnson, Bharath Hariharan, Laurens van~der Maaten, Li~Fei{-}Fei,
  C.~Lawrence Zitnick, and Ross~B. Girshick.
\newblock {CLEVR:} {A} diagnostic dataset for compositional language and
  elementary visual reasoning.
\newblock \emph{CoRR}, abs/1612.06890, 2016.
\newblock URL \url{http://arxiv.org/abs/1612.06890}.

\bibitem[Mostafazadeh et~al.(2016)Mostafazadeh, Misra, Devlin, Zitnick,
  Mitchell, He, and Vanderwende]{vqg}
Nasrin Mostafazadeh, Ishan Misra, Jacob Devlin, Larry Zitnick, Margaret
  Mitchell, Xiaodong He, and Lucy Vanderwende.
\newblock Generating natural questions about an image.
\newblock \emph{CoRR}, abs/1603.06059, 2016.
\newblock URL \url{http://arxiv.org/abs/1603.06059}.

\bibitem[Joshi et~al.(2009)Joshi, Porikli, and Papanikolopoulos]{entropy}
Ajay~J. Joshi, F.~Porikli, and N.~Papanikolopoulos.
\newblock Multi-class active learning for image classification.
\newblock \emph{2009 IEEE Conference on Computer Vision and Pattern
  Recognition}, pages 2372--2379, 2009.

\bibitem[LeCun et~al.(1998)LeCun, Bottou, Bengio, and Haffner]{lenet}
Yann LeCun, Leon Bottou, Yoshua Bengio, and Patrick Haffner.
\newblock lecun-98.pdf.
\newblock \url{http://yann.lecun.com/exdb/publis/pdf/lecun-98.pdf}, 1998.
\newblock (Accessed on 05/30/2021).

\bibitem[Hochreiter and Schmidhuber(1997)]{lstm}
Sepp Hochreiter and J\"{u}rgen Schmidhuber.
\newblock Long short-term memory.
\newblock \emph{Neural Comput.}, 9\penalty0 (8):\penalty0 1735–1780, November
  1997.
\newblock ISSN 0899-7667.
\newblock \doi{10.1162/neco.1997.9.8.1735}.
\newblock URL \url{https://doi.org/10.1162/neco.1997.9.8.1735}.

\bibitem[Huber(1964)]{huber}
Peter~J. Huber.
\newblock {Robust Estimation of a Location Parameter}.
\newblock \emph{The Annals of Mathematical Statistics}, 35\penalty0
  (1):\penalty0 73 -- 101, 1964.
\newblock \doi{10.1214/aoms/1177703732}.
\newblock URL \url{https://doi.org/10.1214/aoms/1177703732}.

\bibitem[Kingma and Ba(2014)]{adam}
Diederik Kingma and Jimmy Ba.
\newblock Adam: A method for stochastic optimization.
\newblock \emph{International Conference on Learning Representations}, 12 2014.

\bibitem[Chevalier-Boisvert et~al.(2018)Chevalier-Boisvert, Willems, and
  Pal]{gym_minigrid}
Maxime Chevalier-Boisvert, Lucas Willems, and Suman Pal.
\newblock Minimalistic gridworld environment for openai gym.
\newblock \url{https://github.com/maximecb/gym-minigrid}, 2018.

\bibitem[Schulman et~al.(2017)Schulman, Wolski, Dhariwal, Radford, and
  Klimov]{ppo}
John Schulman, Filip Wolski, Prafulla Dhariwal, Alec Radford, and Oleg Klimov.
\newblock Proximal policy optimization algorithms, 2017.

\bibitem[Schulman et~al.(2018)Schulman, Moritz, Levine, Jordan, and
  Abbeel]{gae}
John Schulman, Philipp Moritz, Sergey Levine, Michael Jordan, and Pieter
  Abbeel.
\newblock High-dimensional continuous control using generalized advantage
  estimation, 2018.

\bibitem[Bengio et~al.(2009)Bengio, Louradour, Collobert, and
  Weston]{curriculum_learning}
Yoshua Bengio, J\'{e}r\^{o}me Louradour, Ronan Collobert, and Jason Weston.
\newblock Curriculum learning.
\newblock In \emph{Proceedings of the 26th Annual International Conference on
  Machine Learning}, ICML '09, page 41–48, New York, NY, USA, 2009.
  Association for Computing Machinery.
\newblock ISBN 9781605585161.
\newblock \doi{10.1145/1553374.1553380}.
\newblock URL \url{https://doi.org/10.1145/1553374.1553380}.

\bibitem[Perez et~al.(2017)Perez, Strub, de~Vries, Dumoulin, and
  Courville]{film}
Ethan Perez, Florian Strub, Harm de~Vries, Vincent Dumoulin, and Aaron~C.
  Courville.
\newblock Film: Visual reasoning with a general conditioning layer.
\newblock \emph{CoRR}, abs/1709.07871, 2017.
\newblock URL \url{http://arxiv.org/abs/1709.07871}.

\end{thebibliography}

\newpage
\appendix
\section{Appendix}
\subsection{Hyperparameters}

We performed grid search over the hyperparameters set out in Table \ref{tab:hyper} individually for each model.

\begin{table}[h]
\centering
\caption{Hyperparameters: these hyperparameters were selected through grid search using the WandB service.}
\begin{tabular}{lccc}
\hline
\textbf{Agent} & \multicolumn{1}{l}{\textbf{Main}} & \multicolumn{1}{l}{\textbf{FiLM}} & \multicolumn{1}{l}{\textbf{Baseline}} \\ \hline
$\alpha$ & 0.0005 & 0.0001 & 0.001 \\
$\epsilon$ & 0.2 & 0.15 & 0.2 \\
$\gamma$ & 0.99 & 0.99 & 0.99 \\
$\lambda$ & 0.95 & 0.95 & 0.95 \\
$c_1$ & 1 & 1 & 1 \\
$c_2$ & 0.1 & 0.1 & 0.1 \\
$c_3$ & 0.25 & 0.25 & N/A \\
$c_4$ & 1 & 1 & N/A \\
$c_5$ & 0.2 & 0.5 & N/A \\ \hline
\end{tabular}
\label{tab:hyper}
\end{table}

\end{document}